\documentclass[10pt,twocolumn,letterpaper]{article}

\usepackage{cvpr}              
\usepackage{graphicx}
\usepackage{hyperref}       
\usepackage{url}            
\usepackage{booktabs}       
\usepackage{amsfonts}       
\usepackage{nicefrac}       
\usepackage{microtype}      
\usepackage{xcolor}         
\usepackage{amsmath}
\usepackage{epsfig}
\usepackage{graphicx}
\usepackage{amsmath}
\usepackage{subcaption}
\usepackage{witharrows}
\usepackage{rotating}  
\usepackage{multirow}
\usepackage{pifont}
\usepackage{bbding}
\usepackage{adjustbox}
\usepackage{colortbl}
\usepackage{xcolor}
\usepackage{wrapfig}

\definecolor{cvprblue}{rgb}{0.21,0.49,0.74}

\def\paperID{15574} 
\def\confName{CVPR}
\def\confYear{2025}

\title{Reversible Decoupling Network for Single Image Reflection Removal}

\author{Hao Zhao$^{\dag}$, Mingjia Li$^{\dag}$, Qiming Hu, Xiaojie Guo$^{*}$\\
College of Intelligence and Computing, Tianjin University, Tianjin, China\\
{\tt\small xj.max.guo@gmail.com, \{student\_zh, mingjiali, huqiming\}@tju.edu.cn}}

\begin{document}
\maketitle

\footnotetext{\dag The two authors contributed equally. \hspace*{1pt}*Corresponding author.}

\begin{abstract}
Recent deep-learning-based approaches to single-image reflection removal have shown promising advances, primarily for two reasons: 1) the utilization of recognition-pretrained features as inputs, and 2) the design of dual-stream interaction networks. However, according to the Information Bottleneck principle, high-level semantic clues tend to be compressed or discarded during layer-by-layer propagation. Additionally, interactions in dual-stream networks follow a fixed pattern across different layers, limiting overall performance. To address these limitations, we propose a novel architecture called Reversible Decoupling Network (RDNet), which employs a reversible encoder to secure valuable information while flexibly decoupling transmission- and reflection-relevant features during the forward pass. Furthermore, we customize a transmission-rate-aware prompt generator to dynamically calibrate features, further boosting performance. Extensive experiments demonstrate the superiority of RDNet over existing SOTA methods on five widely-adopted benchmark datasets. RDNet achieves the best performance in the NTIRE 2025 Single Image Reflection Removal in the Wild Challenge in both fidelity and perceptual comparison. Our code is available at \href{https://github.com/lime-j/RDNet}{https://github.com/lime-j/RDNet}
\end{abstract}
\section{Introduction}

Reflection is a common superimposition factor when photographing through a transparent medium like glass. In this case, the captured image $I$ contains a mixture of transmission $T$ (the scene behind medium) and reflection $R$ (the reflected scene)~\citep{nayar1997separation}, which can be expressed as $I=T+R$. The presence of reflections often hinders vital information in the transmission layer, impeding the performance of downstream tasks, such as stereo matching, optical flow, and depth estimation \citep{tsin2003stereo,yang2016robust,costanzino2023learning, chen13}. Thus, single-image reflection removal is desired to disentangle the transmission and reflection components from a single input image. However, this is severely ill-posed as infinitely many possible decompositions of $\hat{T}$ and $\hat{R}$ satisfy $I=\hat{T}+\hat{R}$. In other words, it is highly challenging to determine which combination is optimal without effective priors on decomposition. 

\begin{figure}[t]
    \centering
    \includegraphics[width=0.4\textwidth]{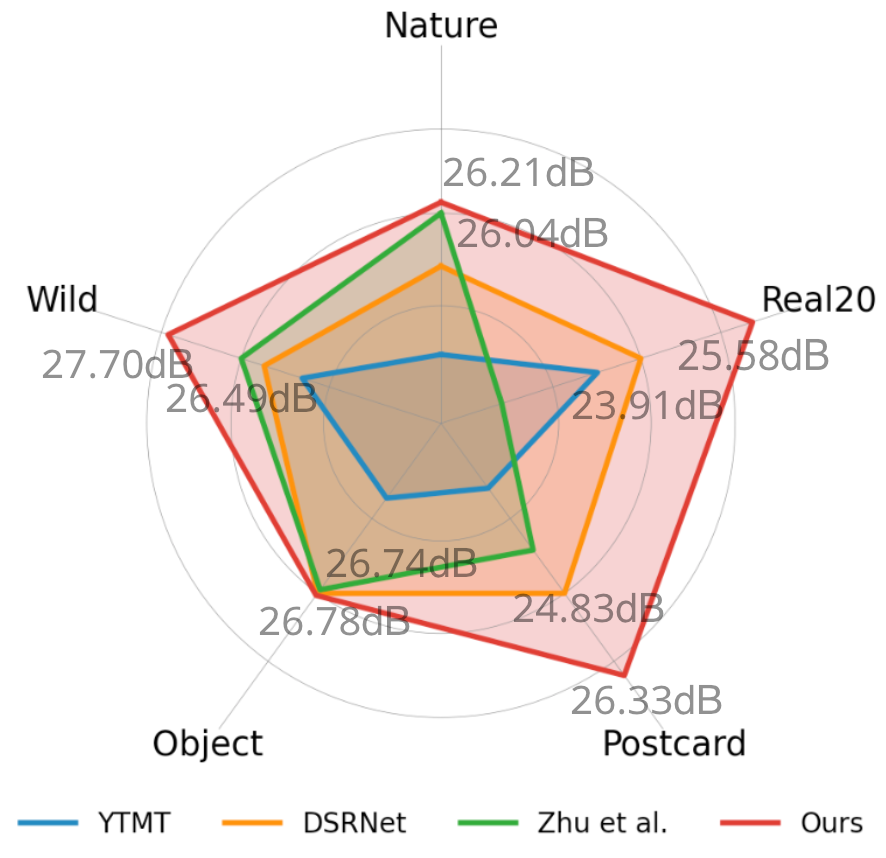} 
    \caption{Quantitative comparison in PSNR between ours and previous state-of-the-art methods, where we achieve new records on all 5 datasets. 
    Note that the scale of each axis is normalized by its second-best value. The best and second-best PSNR values are displayed for reference.}
    \label{fig:radar}
\end{figure}

In recent years, learning-based approaches have made tremendous strides in this field \citep{Zhang_2018_CVPR,Wei_2019_CVPR,apin/LiLYLRZ23,Hu_2023_ICCV,zhong2024language}. A key consensus among these methods is to exploit hierarchical semantic representations through large-scale recognition-pretrained models, which serve as priors/regularizers during training. One pioneering work~\citep{Zhang_2018_CVPR} leverages intermediate features from a pre-trained VGGNet~\citep{iclr/SimonyanZ14a} through the concept of hypercolumns to help differentiate between the transmission and reflection layers from mixtures. Originally from neuroscience, the term ``hypercolumn" refers to a functional unit in the visual cortex that processes visual stimuli at multiple receptive-field sizes~ \citep{hubel1974uniformity}. This concept was first applied to segmentation and localization by interpolating and stacking features extracted from different layers of a network~\citep{Hariharan_2015_CVPR, chen14, chen15}. However, simply mapping stacked high-dimensional hierarchies into a group of much lower-dimensional features--as input for subsequent processes--inevitably leads to considerable semantic information loss.

Previous works~\cite{NEURIPS2021_cf1f78fe,Hu_2023_ICCV} suggest that all information from the source image is valuable for the task. The two components can be optimized by exchanging information between them. For any feasible decomposition ($\hat{T}$, $\hat{R}$), the following relationship holds:
\begin{equation}
\quad \hat{T}:= T-Q,\quad \hat{R}:= R+Q \quad \text{s.t.} \quad I=\hat{T}+\hat{R},
\end{equation}
where $Q$ represents the information to exchange, concretely,  YTMT~\citep{NEURIPS2021_cf1f78fe} and DSRNet~\citep{Hu_2023_ICCV} select $Q$ via activation functions and channel splitting, respectively. Though effective,  the information preservation is not fully guaranteed in their interaction designs, \emph{i.e.}, the information bottleneck induced by linear layers in YTMT, and the multiplicative reductions in the gating mechanism of DSRNet. 

To avoid the above risk, reversible units \citep{DBLP:conf/nips/GomezRUG17}, which are designed to preserve information, may offer a viable solution. In particular, building coupled reversible units naturally fits the situation as follows:
\begin{align}
\underline{\textit{forward:}} 
\begin{array}{c}
    \hat{T}_2 := \hat{T}_1 + \mathcal{F}(\hat{R}_1), \hat{R}_2 := \hat{R}_1 + \mathcal{G}(\hat{T}_2) \\
\end{array}
\\
\underline{\textit{reverse:}}
\begin{array}{c}
    \hat{T}_1 := \hat{T}_2 - \mathcal{F}(\hat{R}_1), \hat{R}_1 := \hat{R}_2 - \mathcal{G}(\hat{T}_2) \\
\end{array}
\end{align}
where $\mathcal{F}(\cdot)$ and $\mathcal{G}(\cdot)$ can be any network modules, and the subscripts stand for the different versions of layer estimations before and after the reversible units, respectively. For simplicity, we also use $\hat{T}$ and $\hat{R}$ to represent the corresponding deep features, which is based on the understanding that if the features are sufficiently disentangled, mapping them back to the image space becomes an easy task.


Although the use of reversible modules can address the issue of information loss in feature interactions at the same scale, preserving multi-scale information during the feedforward process remains a challenge. Beyond the hypercolumn~\citep{Zhang_2018_CVPR} and the progressive hierarchy fusion~\citep{Hu_2023_ICCV}, one intuitive scheme is to stack reversible modules at each scale to facilitate forward propagation while incorporating cross-scale connections to ensure effective multi-scale interaction and fusion. A straightforward approach aligning with this idea is MAXIM~\citep{tu2022maxim} (without consideration of information loss), which employs a fully connected mechanism across multi-scale hierarchies. Similar ideas can also be found in HRNet \citep{sun2019deep}. However, operating on high-dimensional features is computationally expensive and memory-intensive.

Inspired by GLOM \citep{hinton2023represent} and RevCol \citep{revcol}, which employ part-whole hierarchies to represent an image with multiple columns, and embody both bottom-up and top-down interactions to mitigate the computational burden associated with fully connected layers, we integrate multi-scale feature processors into a single sub-network, referred to as a ``column". Further, we ensemble the columns in parallel and build interactions in both bottom-up and top-down manners. It is worth noting that, the scaled residual connections used in GLOM for same-level interactions between adjacent columns can still cause information loss. To remedy this problem, we extend the residual connections by incorporating multi-level reversible connections~\citep{revcol}.


Compared with structural designs guided by information bottleneck principle \citep{tishby2015deep,NEURIPS2021_cf1f78fe}, our proposed framework learns disentangled representations \citep{desjardins2012disentangling,bengio2013representation} by categorizing and recombining the original information, instead of merely selecting and discarding elements, based upon a solid foundation for its information-preserving module (reversible unit). Additionally, it facilitates cross-scale interaction. 
Besides, in real-world scenarios, the reflection pattern varies along with multiple factors, such as the refractive index of the transparent surface, color granularity, and viewing angle \citep{schechner1999polarization}. To enhance the robustness against variations in reflection strength, we further endow the model with an adaptive transmission-rate-aware prompt generator. 

\noindent In light of these considerations, this paper proposes a network, called \textbf{R}eversible \textbf{D}ecoupling \textbf{Net}work (RDNet for short).
The technical contributions of this work are twofold:
\begin{itemize}
\item We revisit the preservation and cross-level interaction problems of hierarchical semantic information during the single image reflection removal/separation. To address the challenges, we introduce a multi-column reversible encoder based on the part-whole hierarchy, complemented by a tailored hierarchy decoder. This design ensures a better retention of rich semantics, effectively mitigating the ill-posed nature of the SIRR task.
  
\item To tackle the varied reflection parameters in real-world scenarios, we introduce an adaptive transmission-rate-aware prompt generator, which learns channel scaling factors during training and leverages this knowledge as a prior when testing. It guides the decomposition network in selecting more accurate transmission-reflection ratios, significantly enhancing the model's generalization capabilities.

\end{itemize}

Extensive experiments are conducted to verify the efficacy of our design, and reveal its superiority over other SOTA alternatives both qualitatively and quantitatively (see Fig.~\ref{fig:radar} for a brief summary).  Notably, our approach also achieves robust generalization on in-the-wild cases, underscoring its practical value in real-world applications (shown by Fig.~\ref{fig:real}).

\section{Related Work}
\label{gen_inst}
\subsection{Single Image Reflection Removal}
\label{sec:mse}
\textbf{Physical formulation.} 
In prevalent reflection removal frameworks~\citep{LrvinPAMI2007}, an image \( I \) is typically decomposed into transmission \( T \) and reflection \( R \) components, so as to \( I = T + R \). However, in real-world scenarios, these two layers may be attenuated by factors such as diffusion and other environmental influences during superposition~\citep{CVPRWAN2020}. To account for such complexities, an augmented modeling has been proposed: \( I=\alpha T + \beta R \), where the coefficients \( \alpha \) and \( \beta \) provide adaptability to varying conditions~\citep{WANCVPR2018,Yang_2018_ECCV}. Nonetheless, the assumption of linear superimposition often breaks down, particularly in cases of overexposure~\citep{Wen_2019_CVPR}. To address this concern, the concept of an alpha-matting map \( W \) is incorporated, leading to a reformulation of the model as \( I = W \circ T + \overline{W} \circ R \) with \( \overline{W} = 1 - W \). While the adjustment improves the model's flexibility, it also increases the complexity of the already ill-posed problem.

The above model struggles to encapsulate the diverse reflection phenomena, highlighting the challenge of developing a universal solution. Hu and Guo~\citep{Hu_2023_ICCV} offered a more comprehensive depiction of the superimposition process by introducing a residual term: \( I = \tilde{T} + \tilde{R} + \phi(T, R) \), where \( \tilde{T} \) and \( \tilde{R} \) signify  the altered transmission and reflection information within \( I \) after superimposition and degradation, as captured by camera sensors. The term \( \phi(T, R)\) denotes the residual information in the reconstruction, arising from factors such as attenuation and overexposure. However, current methods primarily use the above modelings to synthesize training data, expecting the generalizability to real-world data. But, they lack explicit estimation of the physical parameters involved. Furthermore, distance-based loss functions such as mean absolute error (MAE) and mean squared error (MSE) fail to account for global color and intensity shifts. Explicitly estimating the degradation rate of the projected image could improve performance. A more detailed explanation is provided in supplementary.

\noindent\textbf{Deep-learning-based modeling.} 
Zhang \textit{et al.}~\citep{Zhang_2018_CVPR} enhanced semantic awareness by leveraging hypercolumn features extracted from a pre-trained VGG-19 network~\citep{Hariharan_2015_CVPR}, together with perceptual and adversarial losses. ERRNet~\citep{Wei_2019_CVPR} uses misaligned pairs as training data to take a step further. But it overlooks the reflection layer, potentially increasing ambiguity in transmission recovery. Li \textit{et al.}~\citep{apin/LiLYLRZ23} proposed RAGNet, a two-stage network that initially estimates the reflection component and then uses it to guide transmission prediction. Recently, the YTMT strategy proposed in~\citep{NEURIPS2021_cf1f78fe}  treats both components equally through a dual-stream interactive network that restores both layers simultaneously. Yet, noticing the problem hidden in the physical formulation, their interaction module relies on a linear assumption, which may upper-bound its performance. Other methods, such as BDN~\citep{Yang_2018_ECCV} and IBCLN~\citep{Li_2020_CVPR} employ reflection models with scalar weights to iteratively estimate both components, ensuring that the reflection is not too faint. However, the interaction between the two components is ignored, sometimes leading to heavy ghosting effect in transmission and reflection. 
Dong \textit{et al.}~\citep{Dong_2021_ICCV} developed an iterative network that estimates a probabilistic reflection confidence map at each step. DSRNet~\citep{Hu_2023_ICCV} introduces a mutually gated interaction mechanism within a two-stage structural design. In the first stage, the network progressively fuses extracted hierarchical features, while the second stage focuses on further decomposing these features. However, the issue of information loss persists due to the multiplicative reductions in the gating mechanism. Additionally, the progressive hierarchical fusion, isolated in the first stage, does not fully ensure that the hierarchical information is preserved during the subsequent decomposition processes. Zhu \textit{et al.}~\citep{DBLP:journals/corr/abs-2311-17320} proposed a maximum reflection filter for estimating reflection locations and introduce a large dataset, but they similarly overlook interaction between the two layers. Our proposed RDNet addresses the drawbacks of existing approaches by incorporating reversible connections and a multi-column design.

\begin{figure*}[h]
\centering
  \includegraphics[width=\textwidth,height=7.5cm]{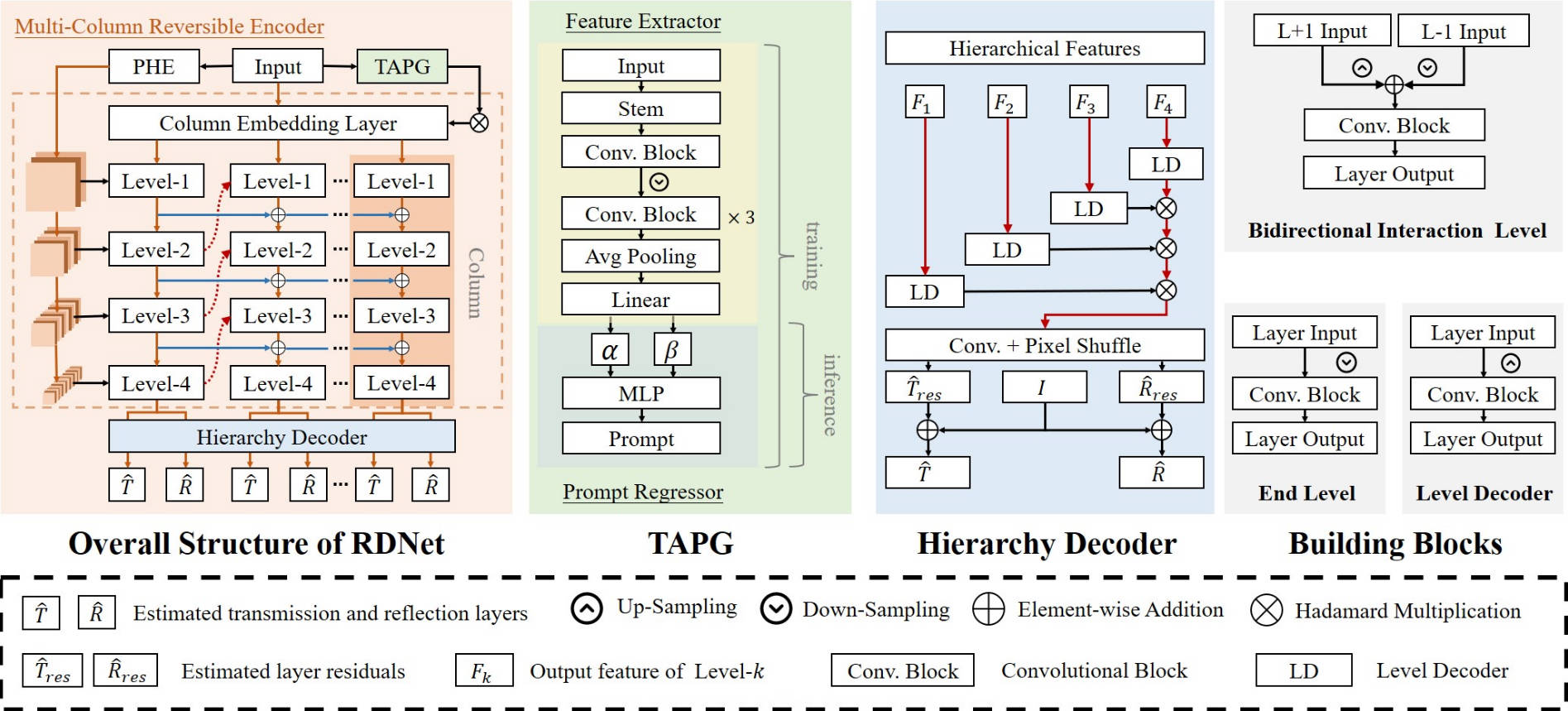}
  \caption{Overall structure of our RDNet, the input is fed in the transmission-rate-aware prompt generator, pretrained hierarchy extractor, and the column embedding. The output of the prompt generator will be transferred into the column network. After interactions between the columns, each column uses a separate decoder to obtain a pair of image layers.}
  \label{fig:netstruc}
  \vspace{-10pt}
\end{figure*}

\subsection{Reversible Network}
Reversible neural networks are designed to prevent information loss by enabling the recovery of original inputs from outputs, thereby maintaining data integrity. Deco and Brauer~\citep{NIPS1994_892c91e0} introduced a reversible architecture that guarantees data preservation through a residual design, which generates a lower triangular Jacobian matrix with unity diagonal elements.  Building upon this concept, Dinh \textit{et al.}~\citep{DBLP:journals/corr/DinhKB14} developed the NICE framework, employing a non-linear bijective transformation between the data and a latent space. However, this design only allows volume-preserving mappings. Dinh \textit{et al.}~\citep{DBLP:conf/iclr/DinhSB17} extended this idea by proposing a reversible transformation that does not require volume preservation. While Gomez \textit{et al.}~\citep{DBLP:conf/nips/GomezRUG17} combined the concept of invertible networks with the ResNet architecture. This manner enables backpropagation without storing the activations in memory, except for a few non-reversible layers.

\noindent\textbf{Reversible Networks for Low-level Vision.} Reversible CNNs have been effectively applied to various low-level tasks, including compression~\citep{DBLP:journals/ijcv/LiuLLYL21},  enhancement~\citep{bmnet, llflow, color} and restoration~\citep{DBLP:journals/tip/HuangD22, rescaling, arb}. These solutions typically employ reversible networks as a shared encoder-decoder in a generative manner, where new textures are generated to supplement lost information during degradation. However, in the task of reflection removal, the target result (the transmission image) is mixed with the reflection rather than lost. This task requires precise decoupling of the input image components instead of generating new textures. To the best of our knowledge, our work is the first to design a reversible architecture specifically for reflection removal.
\noindent\section{Methodology}
In this section, we present the key components of the proposed RDNet, the overall structure of which is schematically depicted in Fig.~\ref{fig:netstruc}. Specifically, it is composed of three primary modules: the multi-column reversible encoder (MCRE), transmission-rate-aware prompt generator (TAPG) and the hierarchy decoder (HDec). The Pretrained Hierarchy Extractor (PHE) captures semantically rich hierarchical representations from the input image and transmits them to each level of the first column in MCRE. Meanwhile, TAPG learns channel-level transmission-reflection ratio priors from the data, mapping these learned fundamental parameters into prompts that guide the MCRE network. Finally, each column in MCRE employs an HDec to encode the hierarchical information, providing effective side guidance~\citep{qin2020u2}. The decoded hierarchies from the last column yield the final results.

\subsection{Multi-scale Reversible Column Encoder}

\label{sec:structure}

As shown in Fig.~\ref{fig:netstruc}, our proposed Multi-Column Reversible Encoder (MCRE), inspired by \citep{revcol}, employs an architecture that differs from end-to-end models \citep{Zhang_2018_CVPR,Wei_2019_CVPR} by incorporating multiple sub-networks, each receiving column embeddings modulated by the Transmission-rate-Aware Prompt Generator (TAPG). The model comprises a Column Embedding Layer and multiple columns encoding multi-scale information. a mIn MCRE, information propagation between columns is handled through two primary mechanisms: intra-level reversible connections (denoted by blue solid lines in the figure) that facilitate information preservation between columns at the same level, and inter-level connections (illustrated as red dashed lines) paired with Bidirectional Interaction Levels, enabling interactions across adjacent levels. This approach effectively decouples multi-scale features up to Level-3. As an exception, Level-4 lacks corresponding cross-level connections, conforming to the structure of the End Level. The initial column within MCRE accepts the hierarchical information extracted by the PHE, ensuring a semantic-rich representation. The subsequent multi-column reversible design ensures the lossless propagation of hierarchical information throughout the decomposition network. 

Specifically, our column embedding layer employs a $7\times7$ convolution layer with a stride of 2, producing $2\times2$ overlapping patches $F_{-1}$ for subsequent processes. Once the embedding is obtained, it is fed to each column.
For the $i$-th column ($i \in \{1, 2, ..., N\} $), each level feature $F_{j}^i, j \in \{0, 1,2\}$ receives information $F_{j-1}^i$ from the lower level of the current column and $F_{j + 1}^{i - 1}$ from a higher level of the previous one. The collected features are further fused with the signal $F_j^{i-1}$ of the current level. The operation described above for the level $j$ is expressed as:
\begin{equation}
    \quad F_j^i = \omega(\theta(F_{j-1}^i)+ \delta(F_{j+1}^{i - 1})) + \gamma F_j^{i-1}.
    \label{eq:forward}
\end{equation}
where $\omega$ denotes the network operation, while $\theta$ and $\delta$ represent downsampling and upsampling operations, respectively. The $\gamma$ term is a simple reversible operation. In our implementation, we utilize a learnable reversible channel-wise scaling as the reversible operation $\gamma$. ~\textbf{This connection is information lossless, as one can retrieve $F_j^{i - 1}$ through the reverse operation:}
\begin{equation}
    \quad F_j^{i - 1} = \gamma^{-1} \left[ F_j^i - \omega(\theta(F_{j-1}^i)+ \delta(F_{j+1}^{i - 1})) \right].
    \label{eq:reverse}
\end{equation}
Notably, for the first level of each column, we define $F_{-1}^i:= F_{-1}$. Moreover, since the last level does not receive any higher-level features, the $\delta(F_{j+1}^{i - 1})$ term is hence discarded.
\begin{figure*}[t]
    \centering 
    \includegraphics[width=\linewidth,height=1.8cm]{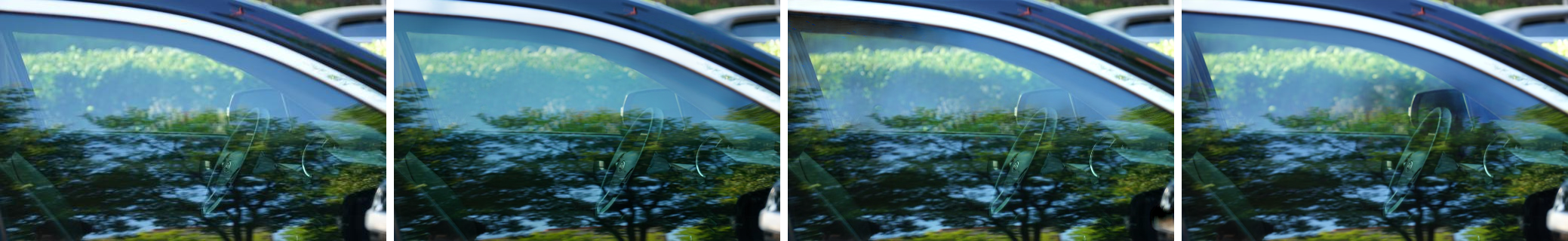}
    \includegraphics[width=\linewidth,height=1.8cm]{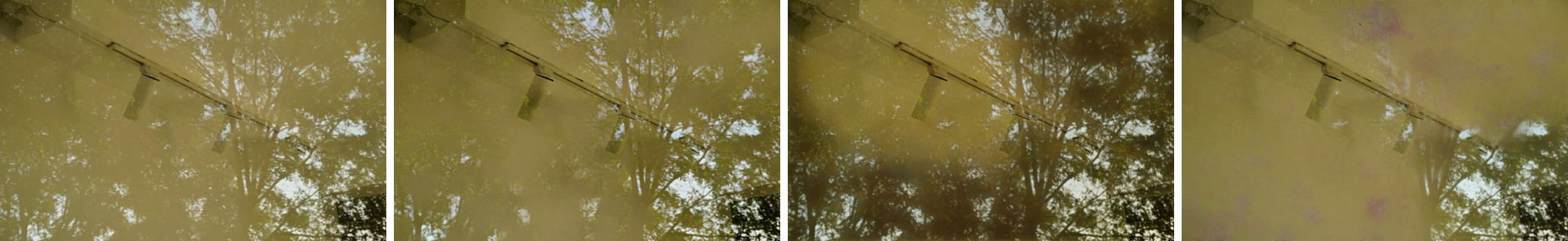}
   \begin{subfigure}{0.24\linewidth}
        \centering
        \subcaption{Input}
        \label{subfig:input}
    \end{subfigure}
    \begin{subfigure}{0.24\linewidth}
        \centering
        \subcaption{ERRNet}
        \label{subfig:ERRNet}
    \end{subfigure}
    \begin{subfigure}{0.24\linewidth}
        \centering
        \subcaption{IBCLN}
        \label{subfig:IBCLN}
    \end{subfigure}
    \begin{subfigure}{0.24\linewidth}
        \centering
        \subcaption{Dong \textit{et al.}}
        \label{subfig:Dong}
    \end{subfigure}
    \includegraphics[width=\linewidth,height=1.8cm]{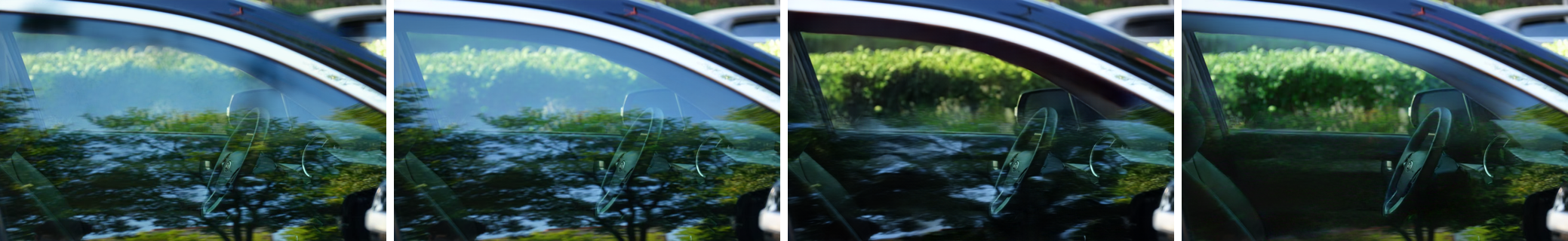}
    \includegraphics[width=\linewidth,height=1.8cm]{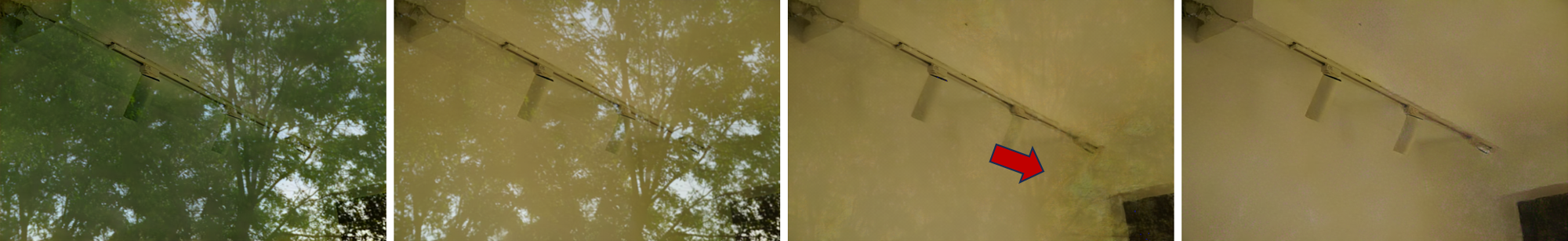}
   \begin{subfigure}{0.24\linewidth}
        \centering
        \subcaption{YTMT}
        \label{subfig:YTMT}
    \end{subfigure}
    \begin{subfigure}{0.24\linewidth}
        \centering
        \subcaption{DSRNet}
        \label{subfig:DSRNet}
    \end{subfigure}
    \begin{subfigure}{0.24\linewidth}
        \centering
        \subcaption{Zhu \textit{et al.}}
        \label{subfig:zhu}
    \end{subfigure}
    \begin{subfigure}{0.24\linewidth}
        \centering
        \subcaption{Ours}
        \label{subfig:input}
    \end{subfigure}
    
    \caption{Qualitative comparisons on  real-world cases. Please zoom in for more details.}
    \label{fig:real}
    \vspace{-10pt}
\end{figure*}

\noindent\textbf{Hierarchy Decoder.} Our hierarchy decoder integrates hierarchical codes from all scales to generate the final output. We leverage several Level Decoders (LD) to interpret the higher-dimensional hierarchies with smaller resolutions into lower-dimensional ones at larger resolutions. The up-sampling operator in an LD is implemented by pixel-shuffle~\citep{DBLP:conf/cvpr/ShiCHTABRW16}, an information-consistent operator before and after the scaling. The up-sampled features are then fused with the information from the previous scale with multiplication modulation. Ultimately, the final LD produces the layer residuals ($\hat{T}_{res}$ and $\hat{R}_{res}$) through another pixel-shuffle up-sampling operation and are connected with the original input to obtain the layer decomposition $\hat{T}$ and $\hat{R}$.

\subsection{Transmission-rate-Aware Prompt Generator}
\label{sec:prompt}

Previous methods for SIRR often exhibit limited generalization capabilities due to the inherent complexity and variability of optical factors in real-world reflective scenarios, compounded by the constraint of limited training data. This limitation can be observed in the real-world test samples we collected, as shown in Fig.~\ref{fig:real}. Meanwhile, in both real-world and synthesized data, color/intensity is often compromised due to the reflection overlaying the transmission, with the transmission $T$ itself being degraded by a transmission rate $a$. In image restoration tasks, the ground truths are typically clean images. However, linearly deviated input/result occurs due to color/illumination shifts in real-world scenarios.

To solve the aforementioned problems, we develop a transmission-rate estimator using a simplified version of the ConvNext model~\citep{Liu_2022_CVPR} pre-trained on ImageNet-1k~\citep{imagenet}. Given an input image $I \in \mathbb{R}^{3\times H \times W} $, our transmission-rate estimator predicts six parameters: $\alpha_{\{R, G, B\}}, \beta_{\{R, G, B\}}$ such that  $\Vert \alpha_{i} T + \beta_{i} - I \Vert_2$ is minimized for each $i \in \{R, G, B\}$. 
When testing the input image using the six parameters generated by the prompt generator, we can obtain an average PSNR of \textbf{24.34dB} across four benchmark datasets (Real20, Objects, Postcard, and Wild), surpassing the previous state-of-the-art method by Dong \textit{et al.}~\citep{Dong_2021_ICCV}. This result confirms the effectiveness of our estimated transmission rate.

Once the transmission rate factor $\alpha_{\{R, G, B\}}, \beta_{\{R, G, B\}}$ is estimated, a three-layer MLP is used to generate prompts for MRCE, resulting in a prompt $P \in \mathbb{R}^{C\times H \times W}$, where $C$ represents the output dimension of the patch embedding layer, set to 64 in our work. Subsequently, the prompt is used to modulate the intermediate features from the column embedding layer $F$ into $P \circ F$, which allows the network to dynamically adapt to the specific characteristics of each input image, thus enhancing the accuracy of reflection removal.

\subsection{Training Objective}
Our model undergoes two training stages. In the first stage, we train the estimator for the transmission rate. Once this is complete, we fix the classifier and proceed to train the main model along with the prompt generator. This training scheme ensures that both the transmission-rate-aware prompt generator and the main model work harmoniously toward the task, resulting in a robust solution.

We employ content loss and perceptual loss for the task, evaluating each pair of images produced by each column using the following loss functions before aggregating them into the final outcome. 
\par \noindent \textbf{Content Loss.} The content loss ensures consistency between the output images and the ground truth training data. In the image domain, we adopt the Mean Squared Error (MSE) loss. Following previous works~\citep{Hu_2023_ICCV, NEURIPS2021_cf1f78fe}, we further regularize the model by encouraging consistency between the output and ground truth in the gradient domain, which writes:
\begin{equation}
\begin{aligned}
    \mathcal{L}_{\text{cont}} := c_0\| \hat{T} - T \|^2_2 + c_1\|\hat{R} - R\|^2_2 
    + c_2\|\nabla \hat{T}-\nabla T\|_{1},
\end{aligned}
\label{eq:contloss}
\end{equation}
where $\|\cdot\|_1$ and $\|\cdot\|_2$ stand for the $\ell_1$ and $\ell_2$ norms, respectively. During the first stage of training, we set $c_0 = 1, c_1 = 0, c2 = 0$. In the second stage, these values are adjusted to $c_0=0.3$, $c_1=0.9$, $c_3=0.6$.

\noindent\textbf{Perceptual Loss.} To enhance the perceptual quality of images produced by our model, we minimize the \(\ell_1\) discrepancy between the features of predicted elements and the ground-truth references. This comparison is made at the `conv2\_2', `conv3\_2', `conv4\_2', and `conv5\_2' layers of a pre-trained VGG-19 network on the ImageNet dataset. Denoting the features at the \(i\)th layer as \(\phi_i(\cdot)\), the perceptual loss is computed as:
\begin{equation}
    \mathcal{L}_{\text{per}} := \sum_j \omega_j \|\phi_j(\hat{T}) - \phi_j(T)\|_1,
\end{equation}
where \(\omega_j\) are weighting coefficients for each layer and \(\omega_j\) $= 0.2$ is empirically set. The total loss turns out to be:
\begin{equation}
    \mathcal{L} := \mathcal{L}_{\text{cont}} + w \mathcal{L}_{\text{per}},
\end{equation}
where $w = 0.01$ is empirically set.

\section{Experimental Validation}
\subsection{Implementation Details}
Our model is implemented in PyTorch~\citep{pytorch} and optimized with Adam optimizer~\citep{ADAM} on an RTX 3090 GPU for 20 epochs. The learning rate is initialized at $10^{-4}$, and remains fixed throughout the training phase, with a batch size of 2. The training dataset comprises both real and synthetic images. To align with previous works, we evaluate the performance of our model under two commonly used data settings: a) The setting from~\citep{NEURIPS2021_cf1f78fe, Wei_2019_CVPR} and~\citep{apin/LiLYLRZ23}, which consists of 90 real image pairs from~\citep{Zhang_2018_CVPR} and 7,643 synthesized pairs from the PASCAL VOC dataset~\citep{PASCol}; and b) The setting from~\citep{Hu_2023_ICCV} and~\citep{Dong_2021_ICCV}, which includes 200 additional real image pairs provided by~\citep{Li_2020_CVPR}. For data synthesizing, we follow the pipeline and physical model from DSRNet~\citep{Hu_2023_ICCV}, represented by $I = \alpha T +  \beta R - T \circ R$. Slightly, we modify this approach by sampling individual $\alpha$ and $\beta$ for R, G, and B channels. This adjustment aims to prevent the transmission rate estimator from converging to a trivial solution. The parameters of the PHE are initialized by a pretrained  FocalNet~\citep{NEURIPS2022_1b08f585}.

\begin{table*}[t]
\centering
\resizebox{\textwidth}{!}{
\begin{tabular}{cccccccccccc}
\toprule
&\multirow{2}{*}{Methods} & \multicolumn{2}{c}{Real20 (20)} & \multicolumn{2}{c}{Objects (200)} & \multicolumn{2}{c}{Postcard (199)} & \multicolumn{2}{c}{Wild (55)} & \multicolumn{2}{c}{\textbf{Average}} \\
\cmidrule(lr){3-4} \cmidrule(lr){5-6} \cmidrule(lr){7-8} \cmidrule(lr){9-10} \cmidrule(lr){11-12}
& & PSNR & SSIM & PSNR & SSIM & PSNR & SSIM & PSNR & SSIM & PSNR & SSIM \\
\midrule
\multirow{7}{*}{\rotatebox{90}{w/o Nat.}} & ERRNet & 22.89 & 0.803 & 24.87 & 0.896 & 22.04 & 0.876 & 24.25 & 0.853 & 23.53 & 0.879 \\
& IBCLN & 21.86 & 0.762 & 24.87 & 0.893 & 23.39 & 0.875 & 24.71 & 0.886 & 24.10 & 0.879 \\
& RAGNet & 22.95 & 0.793 & 26.15 & 0.903 & 23.67 & 0.879 & 25.53 & 0.880 & 24.90 & 0.886 \\
& YTMT & 23.26 & 0.806 & 24.87 & 0.896 & 22.91 & 0.884 & 25.48 & 0.890 & 24.05 & 0.886 \\
& DSRNet & \underline{24.23} & \underline{0.820} & \textbf{26.28} & \textbf{0.914} & \underline{24.56} & \underline{0.908} & \underline{25.68} & \underline{0.896} & \underline{25.40} & \underline{0.905} \\
& Ours & \textbf{24.43} & \textbf{0.835} & \underline{25.76} & \underline{0.905} & \textbf{25.95} & \textbf{0.920} & \textbf{27.20} & \textbf{0.910} & \textbf{25.95} & \textbf{0.908} \\
\midrule
\multirow{4}{*}{\rotatebox{90}{w Nat.}} & Dong \textit{et al.} & 23.34 & 0.812 & 24.36 & 0.898 & 23.72 & 0.903 & 25.73 & 0.902 & 24.21 & 0.897 \\
& DSRNet & \underline{23.91} & \underline{0.818} & \underline{26.74} & 0.920 & \underline{24.83} & \underline{0.911} & 26.11 & 0.906 & \underline{25.75} & 0.910 \\
& Zhu \textit{et al.} & 21.83 & 0.801 & 26.67 & \textbf{0.931} & 24.04 & 0.903 & \underline{26.49} & \underline{0.915} & 25.34 & \underline{0.912} \\
& Ours & \textbf{25.58} & \textbf{0.846} & \textbf{26.78} & \underline{0.921} & \textbf{26.33} & \textbf{0.922} & \textbf{27.70} & \textbf{0.915} & \textbf{26.65} & \textbf{0.917} \\
\bottomrule
\end{tabular}
}
\caption{Quantitative results of various methods on four real-world benchmark datasets. The best results are highlighted in \textbf{bold}, and the second-best results are \underline{underlined}.}
\label{tab:main_comp}
\end{table*}

\subsection{Performance Evaluation}
For the comparison, we evaluate seven state-of-the-art methods: ERRNet~\citep{Wei_2019_CVPR}, IBCLN~\citep{Li_2020_CVPR}, RAGNet~\citep{apin/LiLYLRZ23}, Dong \textit{et al.}~\citep{Dong_2021_ICCV}, YTMT~\citep{NEURIPS2021_cf1f78fe}, DSRNet~\citep{Hu_2023_ICCV}, Zhu \textit{et al.}~\citep{DBLP:journals/corr/abs-2311-17320}, on four real-world datasets, including Real20~\citep{Zhang_2018_CVPR} and three subsets of the $\text{SIR}^2$ Datasets~\citep{Wan_2017_ICCV}, for the Nature~\cite{Dong_2021_ICCV} dataset, we compare IBCLN, ERRNet, YTMT, DSRNet and Zhu \textit{et al.}. 

\begin{table}[t]
\centering
\renewcommand{\arraystretch}{1}
\resizebox{\linewidth}{!}{
\begin{footnotesize}
\begin{tabular}{ccccc}
\toprule
    Setting & Prompt  & Adjusting Input &  PSNR & SSIM \\
    \midrule
    A & $\times$ & $\times$ & 25.52 & 0.909 \\
    B & $\times$ & $\checkmark$ & 25.99 & 0.910 \\ 
    C & $\checkmark$ & $\checkmark$ & 26.03 & 0.913 \\
    Ours & $\checkmark$ & $\times$ & \textbf{26.65} & \textbf{0.917}  \\
    \bottomrule
\end{tabular}
\end{footnotesize}
}
\caption{Ablation studies on the prompt generator. The PSNR and SSIM are calculate across Real20, Objects, Postcard and Wild.}.

\label{tab:ablation_prompt}
\vspace{-10pt}
\end{table}

\begin{table}[t]
\centering
\renewcommand{\arraystretch}{1}
\resizebox{\linewidth}{!}{
\begin{footnotesize}
\begin{tabular}{cccccc}
\toprule
    Setting & Dual-stream & Ref. Loss & Invertibility &  PSNR & SSIM \\
    \midrule
    D & $\checkmark$ & $\checkmark$ & $\checkmark$ & 26.37 & 0.917 \\
    E & $\times$ & $\times$ & $\checkmark$ & 25.99 & 0.914 \\ 
    F & $\times$ & $\checkmark$ & $\times$ & 24.05 & 0.884 \\
    Ours & $\times$ & $\checkmark$ & $\checkmark$ & \textbf{26.65} & \textbf{0.917} \\
    \bottomrule
\end{tabular}
\end{footnotesize}
}
\caption{Ablation studies for network configurations. The PSNR/SSIM are calculated on Real20, Objects, Postcard, and Wild.}
\label{tab:ablation_combined}
\vspace{-5pt}
\end{table}

\begin{table}[t]
\centering
\renewcommand{\arraystretch}{1}
\scalebox{1}{
\begin{footnotesize}
\begin{tabular}{cccccc}
\toprule
    Competitor & DSRNet & Zhu et al.  \\
    \midrule
     Win-rate of our method  & 64.4\% & 77.2\% \\
    \bottomrule
\end{tabular}
\end{footnotesize}
}
\caption{User study comparison with DSRNet and Zhu \textit{et al.}}.
\label{tab:User_study}
\vspace{-23pt}
\end{table}

\noindent\textbf{Quantitative comparisons.} The quantitative result is shown in Tab.~\ref{tab:main_comp}. We directly employ the code and pre-trained weights publicly provided by their authors to obtain all the quantitative results.
To make a fair comparison, the methods with and without additional data from the Nature dataset are compared separately. Apparently, our methods show their superiority over other competitors on all testing datasets, only falling short on SSIM compared to Zhu \textit{et al.} on the Objects dataset. Our methods achieved a promising boost, especially on the Real20 dataset, which contains hard cases collected in real-world conditions, meaning our method can better fit real-world conditions. The other three datasets contain a variety of scenes, illumination conditions, and glass thickness, meaning our method performs better in most conditions. The experimental result demonstrates that our method can adapt to complicated situations and has a stronger generalization ability. For a comprehensive comparison, we present the results obtained on the Nature dataset in supplementary, which comprises 20 real-world samples. Our method achieved the best PSNR and the second-best SSIM, with a marginal decrease of only 0.004 in SSIM. These results further underscore the superiority of our approach in real-world scenarios.
\begin{figure*}[t]
    \centering
    \includegraphics[width=\linewidth,height=2.3cm]{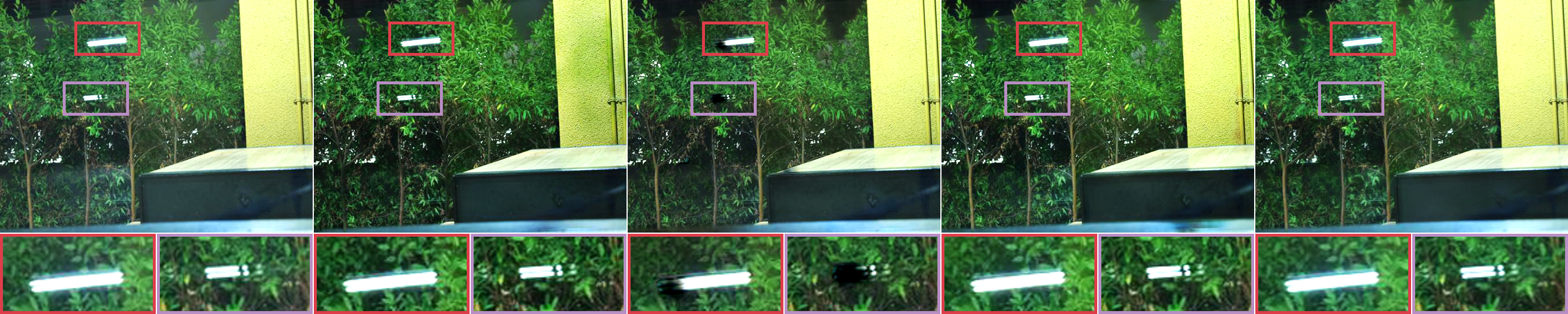} 
    \includegraphics[width=\linewidth,height=2.3cm]{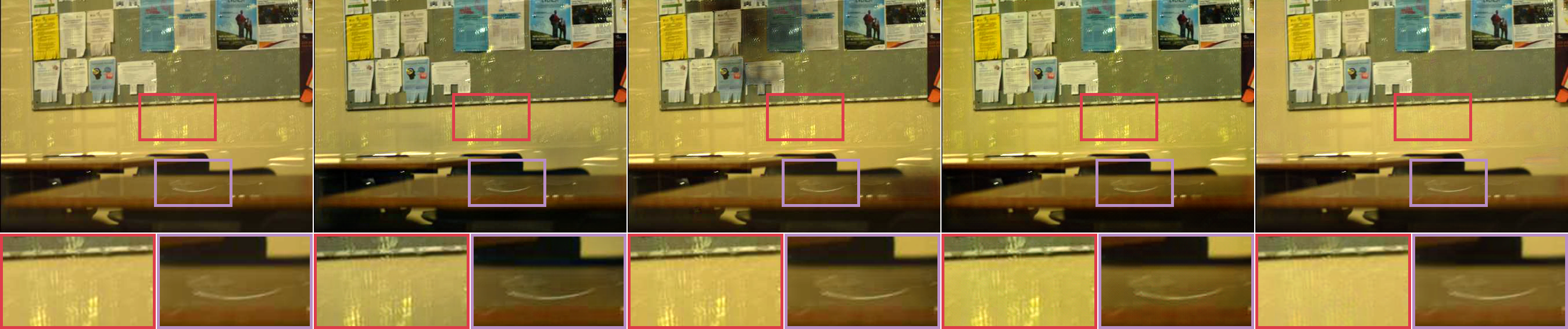}
    \begin{subfigure}{0.19\linewidth}
        \centering
        \subcaption{Input}
        \label{subfig:input}
    \end{subfigure}
    \begin{subfigure}{0.19\linewidth}
        \centering
        \subcaption{ERRNet}
        \label{subfig:ERRNet}
    \end{subfigure}
    \begin{subfigure}{0.19\linewidth}
        \centering
        \subcaption{IBCLN}
        \label{subfig:IBCLN}
    \end{subfigure}
    \begin{subfigure}{0.19\linewidth}
        \centering
        \subcaption{RAGNet}
        \label{subfig:RAGNet}
    \end{subfigure}
    \begin{subfigure}{0.19\linewidth}
        \centering
        \subcaption{Dong \textit{et al.}}
        \label{subfig:Dong}
    \end{subfigure}
    
    \includegraphics[width=\linewidth,height=2.3cm]{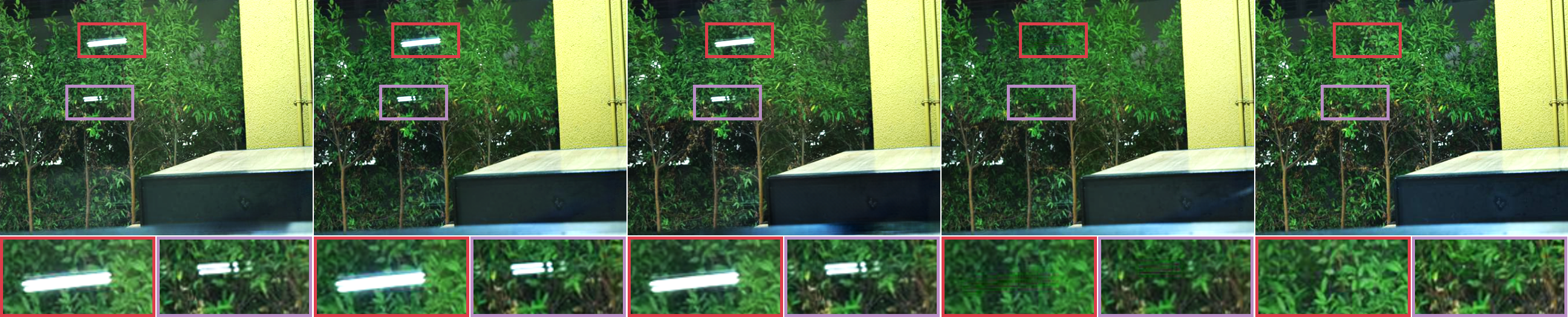}
    \includegraphics[width=\linewidth,height=2.3cm]{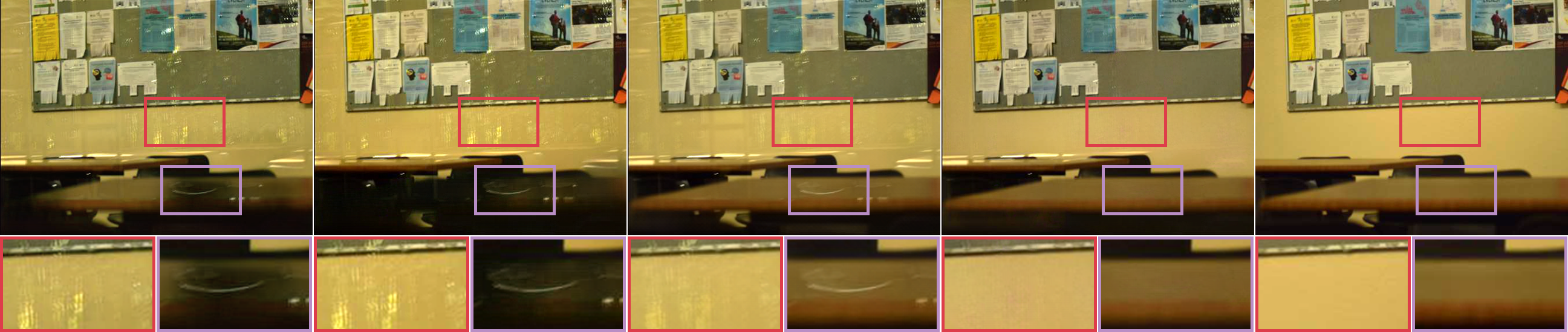}
    \begin{subfigure}{0.19\linewidth}
        \centering
        \subcaption{YTMT}
        \label{subfig:YTMT}
    \end{subfigure}
    \begin{subfigure}{0.19\linewidth}
        \centering
        \subcaption{DSRNet}
        \label{subfig:DSRNet}
    \end{subfigure}
    \begin{subfigure}{0.19\linewidth}
        \centering
        \subcaption{Zhu \textit{et al.}}
        \label{subfig:zhu}
    \end{subfigure}
    \begin{subfigure}{0.19\linewidth}
        \centering
        \subcaption{Ours}
        \label{subfig:input}
    \end{subfigure}
    \begin{subfigure}{0.19\linewidth}
        \centering
        \subcaption{GT}
        \label{subfig:input}
    \end{subfigure}

    \caption{Qualitative comparisons on samples from the Wild dataset. Please zoom in for more details.}
    \label{fig:wild}
    \vspace{-10pt}
\end{figure*}

\noindent\textbf{Qualitative comparisons.} The qualitative comparison is shown in Fig. \ref{fig:wild} and Fig. \ref{fig:real}, with additional examples in the supplementary. Fig. \ref{fig:wild} presents a challenging case with a highly reflective object. Our method effectively removes the reflection, revealing clear texture and color information, outperforming other approaches. In comparison, other methods
struggle to remove the object fully. This example highlights our method's robustness in complex real-world scenarios. The second example further showcases our method's proficiency in handling reflections spread across an image. Here, the reflection is complex and covers a large area, which other methods fail to remove effectively. In contrast, our approach accurately targets and eliminates the majority of the reflection, preserving the integrity of the non-reflective elements. Figure \ref{fig:real} demonstrates the robustness of our method in real-world scenarios, with examples captured under natural conditions. In the first case, a dense reflection covers the car window, posing a significant challenge that most competing methods fail to overcome—only Zhu \textit{et al.} achieves partial removal. In contrast, our approach effectively separates the reflections, producing clearer and more visually appealing results. Similarly, in the second example, our method successfully removes nearly all reflections, while other methods struggle to perform effectively. These cases underscore the robustness of our decoupling approach.

\noindent\textbf{User Study.} We present our user study in Tab. \ref{tab:User_study}. Inspired by LIME-Eval\cite{DBLP:journals/corr/abs-2410-08810}, we asked 20 users to choose their favorable one from each pair of outputs. All five datasets are included and 587 pairs of options are obtained. Our method achieves a win rate of 77.2\% to Zhu et al., and 64.4\% to DSRNet, indicating better performance.


\subsection{Ablation Studies}
To better verify the effect of our prompt generator and reversible structure, we provide a series of ablation studies for the key component of our design as below.

\noindent\textbf{Discussion on transmission-rate-aware prompt generator.}
To inform the model with the transmission rate, a straightforward approach is to adjust the input image to enhance it globally using the estimated transmission rate. Specifically, for \( I := a T + b R + \phi(T, R) \), we adjust the input \( I \) to \(\frac{1}{a} I := T + \frac{b}{a} R + \frac{1}{a} \phi(T, R) \). This operation is denoted as Adjusting Input in Tab.~\ref{tab:ablation_prompt}. As shown in Tab.~\ref{tab:ablation_prompt}, if we remove all transmission-rate-aware techniques (setting A), the average performance drops by 1.13 dB. If we adopt the straightforward method described above (setting B), the performance recovers by 0.47 dB. This confirms the importance of informing the model with the transmission rate.  However, as we analyzed in Section~\ref{sec:prompt}, directly adjusting the input image is far from optimal. Due to potential inaccuracies in the estimation in some scenarios, directly adjusting the model can introduce an additional shift that is difficult to correct during second-stage training. A more subtle and flexible approach is to reweight the feature channels with our transmission-rate-aware prompt. To verify this, we both adjust the input and add a transmission-rate-aware prompt to the feature (setting C). The performance remains nearly the same as in Setting B, indicating that adjusting the input makes it challenging for the model to recover from incorrect estimations. Finally, our model with the proposed transmission-rate-aware prompt outperforms all variants, demonstrating its efficacy.

\noindent\textbf{Discussion on model design.}
To verify the rationality of our design of the decoupling model, we created three new variants of our model. We modify our RDNet to a DSRNet-style one, where two streams estimate transmission and reflection separately in a single column and interact with each other (Setting D). As shown in Tab. \ref{tab:ablation_combined}, even with double computation, the performance still drops by 0.28 dB. This confirms the effectiveness of the decoupling design compared to the dual-stream design. Secondly, we removed the reflection part ($c_1\|\hat{R} - R\|^2_2 $) in the content loss function (Eq.~\ref{eq:contloss}), leaving only the transmission part ($c_0\| \hat{T} - T \|^2_2 + c_2\|\nabla \hat{T}-\nabla T\|_{1}$) in the training process. This variant is denoted as Setting E. A performance drop of 0.66dB can be observed. This confirms the necessity of the reflection loss function. Without regularization predicting the other component, the network weakens its ability to clearly identify both components in single-stream feature maps. To verify the necessity of the invertibility of the network, we replace the reversible connection with the U-Net connection~\cite{Unet} (Setting F). With slightly more parameters and much more memory, a massive performance drop of 2.6 dB can be discovered, 
indicating the importance of invertibility. 

\begin{table}[t]
    \centering
    \renewcommand{\arraystretch}{1}
    \scalebox{1}{
        \begin{tabular}{cccc}
            \toprule
            {Column count}   & 2     & 4     & 6     \\
            \midrule
            PSNR & 26.25 & \textbf{26.65} & 26.19 \\
             SSIM & 0.914 & \textbf{0.917} & 0.910 \\
            \midrule
        \end{tabular}
    }
    \caption{The experiment of changing numbers of columns.}
    \label{tab:colabla}
    \vspace{-10pt}
\end{table}

\noindent\textbf{Discussion on the number of columns.} We investigate the effect of the number of columns in Tab.\ref{tab:colabla}. Specifically, we adjusted the number of columns with configurations of 2, 4, and 6 columns. Our findings indicate that our choice of 4 columns yields the highest performance. In contrast, configurations with 2 and 6 columns resulted in performance drops of 0.4dB and 0.46dB in PSNR, respectively. This suggests that an optimal balance exists, where too few or too many columns can detract from the model's performance.
\section{Conclusion}
In this paper, we proposed RDNet, a novel model for addressing key challenges in the task of single-image reflection removal. Specifically, RDNet tackles the limitations of insufficient utilization of multi-scale, pretrained hierarchical information and information loss during feature decoupling. The multi-column reversible structure enables the preservation of rich semantic features, which are then effectively leveraged in the multi-scale processing of each column. Furthermore, the proposed Transmission-rate-Aware Prompt Generator alleviates the inherent conflict between complex reflection parameters and limited training data. Through these innovations, RDNet demonstrates an enhanced capability for robust reflection removal. Our method demonstrates superior performance compared to state-of-the-art techniques across real-world benchmark datasets, highlighting its robustness and adaptability in diverse reflective scenarios. It is positive that our work opens up new avenues for research in reflection removal and has the potential to significantly impact various applications in image restoration.

\noindent \textbf{Acknowledgement.}
This work was supported by the National Natural Science Foundation of China under Grant nos.62372251 and 62072327. The computational resources of this work is partially supported by TPU Research Cloud (TRC) and DataCanvas Limited.
{
    \small
    \bibliographystyle{ieeenat_fullname}
    \bibliography{main}

\begin{thebibliography}{54}
\providecommand{\natexlab}[1]{#1}
\providecommand{\url}[1]{\texttt{#1}}
\expandafter\ifx\csname urlstyle\endcsname\relax
  \providecommand{\doi}[1]{doi: #1}\else
  \providecommand{\doi}{doi: \begingroup \urlstyle{rm}\Url}\fi

\bibitem[Bengio et~al.(2013)Bengio, Courville, and Vincent]{bengio2013representation}
Yoshua Bengio, Aaron Courville, and Pascal Vincent.
\newblock Representation learning: A review and new perspectives.
\newblock \emph{TPAMI}, 35\penalty0 (8):\penalty0 1798--1828, 2013.

\bibitem[Cai et~al.(2023)Cai, Zhou, Han, Sun, Kong, Li, and Zhang]{revcol}
Yuxuan Cai, Yizhuang Zhou, Qi Han, Jianjian Sun, Xiangwen Kong, Jun Li, and Xiangyu Zhang.
\newblock Reversible column networks.
\newblock In \emph{ICLR}, 2023.

\bibitem[Costanzino et~al.(2023)Costanzino, Ramirez, Poggi, Tosi, Mattoccia, and Di~Stefano]{costanzino2023learning}
Alex Costanzino, Pierluigi~Zama Ramirez, Matteo Poggi, Fabio Tosi, Stefano Mattoccia, and Luigi Di~Stefano.
\newblock Learning depth estimation for transparent and mirror surfaces.
\newblock In \emph{ICCV}, pages 9244--9255, 2023.

\bibitem[Deco and Brauer(1994)]{NIPS1994_892c91e0}
Gustavo Deco and Wilfried Brauer.
\newblock Higher order statistical decorrelation without information loss.
\newblock In \emph{NeurIPS}, 1994.

\bibitem[Deng et~al.(2009)Deng, Dong, Socher, Li, Li, and Fei-Fei]{imagenet}
Jia Deng, Wei Dong, Richard Socher, Li-Jia Li, Kai Li, and Li Fei-Fei.
\newblock Imagenet: A large-scale hierarchical image database.
\newblock pages 248--255, 2009.

\bibitem[Desjardins et~al.(2012)Desjardins, Courville, and Bengio]{desjardins2012disentangling}
Guillaume Desjardins, Aaron Courville, and Yoshua Bengio.
\newblock Disentangling factors of variation via generative entangling.
\newblock \emph{arXiv preprint}, 2012.

\bibitem[Dinh et~al.(2015)Dinh, Krueger, and Bengio]{DBLP:journals/corr/DinhKB14}
Laurent Dinh, David Krueger, and Yoshua Bengio.
\newblock {NICE:} non-linear independent components estimation.
\newblock In \emph{ICLR}, 2015.

\bibitem[Dinh et~al.(2017)Dinh, Sohl{-}Dickstein, and Bengio]{DBLP:conf/iclr/DinhSB17}
Laurent Dinh, Jascha Sohl{-}Dickstein, and Samy Bengio.
\newblock Density estimation using real {NVP}.
\newblock In \emph{ICLR}, 2017.

\bibitem[Dong et~al.(2021)Dong, Xu, Yang, Bao, Xu, and Lau]{Dong_2021_ICCV}
Zheng Dong, Ke Xu, Yin Yang, Hujun Bao, Weiwei Xu, and Rynson~W.H. Lau.
\newblock Location-aware single image reflection removal.
\newblock In \emph{ICCV}, 2021.

\bibitem[Everingham et~al.(2010)Everingham, Gool, Williams, Winn, and Zisserman]{PASCol}
Mark Everingham, Luc~Van Gool, Christopher K.~I. Williams, John~M. Winn, and Andrew Zisserman.
\newblock The pascal visual object classes {(VOC)} challenge.
\newblock \emph{IJCV}, 88\penalty0 (2):\penalty0 303--338, 2010.

\bibitem[Gomez et~al.(2017)Gomez, Ren, Urtasun, and Grosse]{DBLP:conf/nips/GomezRUG17}
Aidan~N. Gomez, Mengye Ren, Raquel Urtasun, and Roger~B. Grosse.
\newblock The reversible residual network: Backpropagation without storing activations.
\newblock In \emph{NeurIPS}, pages 2214--2224, 2017.

\bibitem[Hariharan et~al.(2015)Hariharan, Arbelaez, Girshick, and Malik]{Hariharan_2015_CVPR}
Bharath Hariharan, Pablo Arbelaez, Ross Girshick, and Jitendra Malik.
\newblock Hypercolumns for object segmentation and fine-grained localization.
\newblock In \emph{CVPR}, 2015.

\bibitem[Hinton(2023)]{hinton2023represent}
Geoffrey Hinton.
\newblock How to represent part-whole hierarchies in a neural network.
\newblock \emph{Neural Computation}, 35\penalty0 (3):\penalty0 413--452, 2023.

\bibitem[Hu and Guo(2021)]{NEURIPS2021_cf1f78fe}
Qiming Hu and Xiaojie Guo.
\newblock Trash or treasure? an interactive dual-stream strategy for single image reflection separation.
\newblock In \emph{NeurIPS}, 2021.

\bibitem[Hu and Guo(2023)]{Hu_2023_ICCV}
Qiming Hu and Xiaojie Guo.
\newblock Single image reflection separation via component synergy.
\newblock In \emph{ICCV}, 2023.

\bibitem[Huang and Dragotti(2022)]{DBLP:journals/tip/HuangD22}
Junjie Huang and Pier~Luigi Dragotti.
\newblock Winnet: Wavelet-inspired invertible network for image denoising.
\newblock \emph{IEEE TIP}, 31:\penalty0 4377--4392, 2022.

\bibitem[Hubel and Wiesel(1974)]{hubel1974uniformity}
David~H Hubel and Torsten~N Wiesel.
\newblock Uniformity of monkey striate cortex: a parallel relationship between field size, scatter, and magnification factor.
\newblock \emph{Journal of Comparative Neurology}, 158\penalty0 (3):\penalty0 295--305, 1974.

\bibitem[Jiang et~al.(2024)Jiang, Chen, Pun, Wang, and Feng]{chen13}
Yiguo Jiang, Xuhang Chen, Chi-Man Pun, Shuqiang Wang, and Wei Feng.
\newblock Mfdnet: Multi-frequency deflare network for efficient nighttime flare removal.
\newblock \emph{The Visual Computer}, 40\penalty0 (11):\penalty0 7575--7588, 2024.

\bibitem[Kingma and Ba(2015)]{ADAM}
Diederik~P. Kingma and Jimmy Ba.
\newblock Adam: {A} method for stochastic optimization.
\newblock In \emph{ICLR}, 2015.

\bibitem[Levin and Weiss(2007)]{LrvinPAMI2007}
Anat Levin and Yair Weiss.
\newblock User assisted separation of reflections from a single image using a sparsity prior.
\newblock \emph{IEEE TPAMI}, 29\penalty0 (9):\penalty0 1647--1654, 2007.

\bibitem[Li et~al.(2020)Li, Yang, He, Lin, and Hopcroft]{Li_2020_CVPR}
Chao Li, Yixiao Yang, Kun He, Stephen Lin, and John~E. Hopcroft.
\newblock Single image reflection removal through cascaded refinement.
\newblock In \emph{CVPR}, 2020.

\bibitem[Li et~al.(2024)Li, Zhao, and Guo]{DBLP:journals/corr/abs-2410-08810}
Mingjia Li, Hao Zhao, and Xiaojie Guo.
\newblock Lime-eval: Rethinking low-light image enhancement evaluation via object detection.
\newblock \emph{CoRR}, abs/2410.08810, 2024.

\bibitem[Li et~al.(2022)Li, Lien, and Wang]{color}
Yuan{-}kui Li, Yun{-}Hsuan Lien, and Yu{-}Shuen Wang.
\newblock Style-structure disentangled features and normalizing flows for diverse icon colorization.
\newblock In \emph{CVPR}, pages 11234--11243, 2022.

\bibitem[Li et~al.(2023)Li, Liu, Yi, Li, Ren, and Zuo]{apin/LiLYLRZ23}
Yu Li, Ming Liu, Yaling Yi, Qince Li, Dongwei Ren, and Wangmeng Zuo.
\newblock Two-stage single image reflection removal with reflection-aware guidance.
\newblock \emph{Applied Intelligence}, 53\penalty0 (16):\penalty0 19433--19448, 2023.

\bibitem[Liu et~al.(2021)Liu, Liu, Li, Yan, and Li]{DBLP:journals/ijcv/LiuLLYL21}
Kang Liu, Dong Liu, Li Li, Ning Yan, and Houqiang Li.
\newblock Semantics-to-signal scalable image compression with learned revertible representations.
\newblock \emph{IJCV}, 129\penalty0 (9):\penalty0 2605--2621, 2021.

\bibitem[Liu et~al.(2024)Liu, Huang, Yuan, Zheng, Zhong, Chen, and Pun]{chen15}
Xinyi Liu, Guoheng Huang, Xiaochen Yuan, Zewen Zheng, Guo Zhong, Xuhang Chen, and Chi-Man Pun.
\newblock Weakly supervised semantic segmentation via saliency perception with uncertainty-guided noise suppression.
\newblock \emph{The Visual Computer}, pages 1--16, 2024.

\bibitem[Liu et~al.(2022)Liu, Mao, Wu, Feichtenhofer, Darrell, and Xie]{Liu_2022_CVPR}
Zhuang Liu, Hanzi Mao, Chao-Yuan Wu, Christoph Feichtenhofer, Trevor Darrell, and Saining Xie.
\newblock A convnet for the 2020s.
\newblock In \emph{CVPR}, 2022.

\bibitem[Nayar et~al.(1997)Nayar, Fang, and Boult]{nayar1997separation}
Shree~K Nayar, Xi-Sheng Fang, and Terrance Boult.
\newblock Separation of reflection components using color and polarization.
\newblock \emph{IJCV}, 21\penalty0 (3):\penalty0 163--186, 1997.

\bibitem[Paszke et~al.(2019)Paszke, Gross, Massa, Lerer, Bradbury, Chanan, Killeen, Lin, Gimelshein, Antiga, Desmaison, Kopf, Yang, DeVito, Raison, Tejani, Chilamkurthy, Steiner, Fang, Bai, and Chintala]{pytorch}
Adam Paszke, Sam Gross, Francisco Massa, Adam Lerer, James Bradbury, Gregory Chanan, Trevor Killeen, Zeming Lin, Natalia Gimelshein, Luca Antiga, Alban Desmaison, Andreas Kopf, Edward Yang, Zachary DeVito, Martin Raison, Alykhan Tejani, Sasank Chilamkurthy, Benoit Steiner, Lu Fang, Junjie Bai, and Soumith Chintala.
\newblock Pytorch: An imperative style, high-performance deep learning library.
\newblock In \emph{NeurIPS}, 2019.

\bibitem[Qin et~al.(2020)Qin, Zhang, Huang, Dehghan, Zaiane, and Jagersand]{qin2020u2}
Xuebin Qin, Zichen Zhang, Chenyang Huang, Masood Dehghan, Osmar~R Zaiane, and Martin Jagersand.
\newblock U2-net: Going deeper with nested u-structure for salient object detection.
\newblock \emph{Pattern recognition}, 106:\penalty0 107404, 2020.

\bibitem[Ronneberger et~al.(2015)Ronneberger, Fischer, and Brox]{Unet}
Olaf Ronneberger, Philipp Fischer, and Thomas Brox.
\newblock U-net: Convolutional networks for biomedical image segmentation.
\newblock In \emph{MICCAI}, 2015.

\bibitem[Schechner et~al.(1999)Schechner, Shamir, and Kiryati]{schechner1999polarization}
Yoav~Y Schechner, Joseph Shamir, and Nahum Kiryati.
\newblock Polarization-based decorrelation of transparent layers: The inclination angle of an invisible surface.
\newblock In \emph{ICCV}, pages 814--819, 1999.

\bibitem[Shi et~al.(2016)Shi, Caballero, Huszar, Totz, Aitken, Bishop, Rueckert, and Wang]{DBLP:conf/cvpr/ShiCHTABRW16}
Wenzhe Shi, Jose Caballero, Ferenc Huszar, Johannes Totz, Andrew~P. Aitken, Rob Bishop, Daniel Rueckert, and Zehan Wang.
\newblock Real-time single image and video super-resolution using an efficient sub-pixel convolutional neural network.
\newblock In \emph{CVPR}, pages 1874--1883, 2016.

\bibitem[Simonyan and Zisserman(2015)]{iclr/SimonyanZ14a}
Karen Simonyan and Andrew Zisserman.
\newblock Very deep convolutional networks for large-scale image recognition.
\newblock In \emph{ICLR}, 2015.

\bibitem[Sun et~al.(2019)Sun, Xiao, Liu, and Wang]{sun2019deep}
Ke Sun, Bin Xiao, Dong Liu, and Jingdong Wang.
\newblock Deep high-resolution representation learning for human pose estimation.
\newblock In \emph{CVPR}, pages 5693--5703, 2019.

\bibitem[Tishby and Zaslavsky(2015)]{tishby2015deep}
Naftali Tishby and Noga Zaslavsky.
\newblock Deep learning and the information bottleneck principle.
\newblock In \emph{IEEE Information Theory Workshop}, pages 1--5. IEEE, 2015.

\bibitem[Tsin et~al.(2003)Tsin, Kang, and Szeliski]{tsin2003stereo}
Yanghai Tsin, Sing~Bing Kang, and Richard Szeliski.
\newblock Stereo matching with reflections and translucency.
\newblock In \emph{CVPR}, pages I--I, 2003.

\bibitem[Tu et~al.(2022)Tu, Talebi, Zhang, Yang, Milanfar, Bovik, and Li]{tu2022maxim}
Zhengzhong Tu, Hossein Talebi, Han Zhang, Feng Yang, Peyman Milanfar, Alan Bovik, and Yinxiao Li.
\newblock Maxim: Multi-axis mlp for image processing.
\newblock In \emph{CVPR}, pages 5769--5780, 2022.

\bibitem[Wan et~al.(2017)Wan, Shi, Duan, Tan, and Kot]{Wan_2017_ICCV}
Renjie Wan, Boxin Shi, Ling-Yu Duan, Ah-Hwee Tan, and Alex~C. Kot.
\newblock Benchmarking single-image reflection removal algorithms.
\newblock In \emph{ICCV}, 2017.

\bibitem[Wan et~al.(2018)Wan, Shi, Duan, Tan, and Kot]{WANCVPR2018}
Renjie Wan, Boxin Shi, Ling{-}Yu Duan, Ah{-}Hwee Tan, and Alex~C. Kot.
\newblock {CRRN:} multi-scale guided concurrent reflection removal network.
\newblock In \emph{CVPR}, 2018.

\bibitem[Wan et~al.(2020)Wan, Shi, Li, Duan, and Kot]{CVPRWAN2020}
Renjie Wan, Boxin Shi, Haoliang Li, Ling{-}Yu Duan, and Alex~C. Kot.
\newblock Reflection scene separation from a single image.
\newblock In \emph{CVPR}, 2020.

\bibitem[Wang et~al.(2022)Wang, Wan, Yang, Li, Chau, and Kot]{llflow}
Yufei Wang, Renjie Wan, Wenhan Yang, Haoliang Li, Lap{-}Pui Chau, and Alex~C. Kot.
\newblock Low-light image enhancement with normalizing flow.
\newblock In \emph{AAAI}, pages 2604--2612, 2022.

\bibitem[Wei et~al.(2019)Wei, Yang, Fu, Wipf, and Huang]{Wei_2019_CVPR}
Kaixuan Wei, Jiaolong Yang, Ying Fu, David Wipf, and Hua Huang.
\newblock Single image reflection removal exploiting misaligned training data and network enhancements.
\newblock In \emph{CVPR}, 2019.

\bibitem[Wen et~al.(2019)Wen, Tan, Qin, Liu, Han, and He]{Wen_2019_CVPR}
Qiang Wen, Yinjie Tan, Jing Qin, Wenxi Liu, Guoqiang Han, and Shengfeng He.
\newblock Single image reflection removal beyond linearity.
\newblock In \emph{CVPR}, 2019.

\bibitem[Yang et~al.(2016)Yang, Li, Dai, and Tan]{yang2016robust}
Jiaolong Yang, Hongdong Li, Yuchao Dai, and Robby~T Tan.
\newblock Robust optical flow estimation of double-layer images under transparency or reflection.
\newblock In \emph{CVPR}, pages 1410--1419, 2016.

\bibitem[Yang et~al.(2018)Yang, Gong, Liu, and Shi]{Yang_2018_ECCV}
Jie Yang, Dong Gong, Lingqiao Liu, and Qinfeng Shi.
\newblock Seeing deeply and bidirectionally: A deep learning approach for single image reflection removal.
\newblock In \emph{ECCV}, 2018.

\bibitem[Yang et~al.(2022)Yang, Li, Dai, and Gao]{NEURIPS2022_1b08f585}
Jianwei Yang, Chunyuan Li, Xiyang Dai, and Jianfeng Gao.
\newblock Focal modulation networks.
\newblock In \emph{NeurIPS}, 2022.

\bibitem[Yao et~al.(2023)Yao, Tsao, Lo, Tseng, Chang, and Lee]{arb}
Jie{-}En Yao, Li{-}Yuan Tsao, Yi{-}Chen Lo, Roy Tseng, Chia{-}Che Chang, and Chun{-}Yi Lee.
\newblock Local implicit normalizing flow for arbitrary-scale image super-resolution.
\newblock In \emph{CVPR}, pages 1776--1785, 2023.

\bibitem[Zhang et~al.(2018)Zhang, Ng, and Chen]{Zhang_2018_CVPR}
Xuaner Zhang, Ren Ng, and Qifeng Chen.
\newblock Single image reflection separation with perceptual losses.
\newblock In \emph{CVPR}, 2018.

\bibitem[Zheng et~al.(2024)Zheng, Chen, Liu, Li, Lei, He, Pun, and Zhou]{chen14}
Fuchen Zheng, Xuhang Chen, Weihuang Liu, Haolun Li, Yingtie Lei, Jiahui He, Chi-Man Pun, and Shounjun Zhou.
\newblock Smaformer: Synergistic multi-attention transformer for medical image segmentation.
\newblock In \emph{BIBM}, pages 4048--4053, 2024.

\bibitem[Zhong et~al.(2024)Zhong, Hong, Weng, Liang, and Shi]{zhong2024language}
Haofeng Zhong, Yuchen Hong, Shuchen Weng, Jinxiu Liang, and Boxin Shi.
\newblock Language-guided image reflection separation.
\newblock 2024.

\bibitem[Zhu et~al.(2022)Zhu, Huang, Fu, Zhao, Sun, and Zha]{bmnet}
Yurui Zhu, Jie Huang, Xueyang Fu, Feng Zhao, Qibin Sun, and Zheng{-}Jun Zha.
\newblock Bijective mapping network for shadow removal.
\newblock In \emph{CVPR}, pages 5617--5626. {IEEE}, 2022.

\bibitem[Zhu et~al.(2023)Zhu, Wang, Dong, Zhang, Gao, and Yuan]{rescaling}
Yiming Zhu, Cairong Wang, Chenyu Dong, Ke Zhang, Hongyang Gao, and Chun Yuan.
\newblock High-frequency normalizing flow for image rescaling.
\newblock \emph{{IEEE} TIP}, 32:\penalty0 6223--6233, 2023.

\bibitem[Zhu et~al.(2024)Zhu, Fu, Jiang, Zhang, Sun, Chen, Zha, and Li]{DBLP:journals/corr/abs-2311-17320}
Yurui Zhu, Xueyang Fu, Peng{-}Tao Jiang, Hao Zhang, Qibin Sun, Jinwei Chen, Zheng{-}Jun Zha, and Bo Li.
\newblock Revisiting single image reflection removal in the wild.
\newblock In \emph{CVPR}, 2024.

\end{thebibliography}
}


\end{document}